\pdfoutput=1

\documentclass[11pt]{article}

\usepackage{amsmath}
\usepackage{amssymb}
\usepackage{mathtools}
\usepackage{amsthm}
\usepackage{enumitem}
\usepackage{algorithm}
\usepackage{algorithmic}
\usepackage{multicol}

\usepackage[final]{acl}

\usepackage{times}
\usepackage{latexsym}

\usepackage[T1]{fontenc}

\usepackage[utf8]{inputenc}

\usepackage{microtype}

\usepackage{inconsolata}

\usepackage{arydshln}
\usepackage{wrapfig}
\usepackage{multirow}
\usepackage{tabularx}
\usepackage{enumitem}
\usepackage{subcaption}
\usepackage{svg}
\usepackage{caption}
\usepackage{booktabs}
\usepackage{colortbl}
\usepackage{bbding}
\usepackage{float}
\usepackage{tabularray} 
\usepackage{physics}

\usepackage{graphicx}
\newcommand{\beq}{\vspace{0mm}\begin{equation}}
\newcommand{\eeq}{\vspace{0mm}\end{equation}}
\newcommand{\beqs}{\vspace{0mm}\begin{eqnarray}}
\newcommand{\eeqs}{\vspace{0mm}\end{eqnarray}}
\newcommand{\barr}{\begin{array}}
\newcommand{\earr}{\end{array}}

\usepackage{booktabs}
\usepackage{array}
\newcolumntype{L}[1]{>{\raggedright\let\newline\\\arraybackslash\hspace{0pt}}m{#1}}
\newcolumntype{C}[1]{>{\centering\let\newline\\\arraybackslash\hspace{0pt}}m{#1}}
\newcolumntype{R}[1]{>{\raggedleft\let\newline\\\arraybackslash\hspace{0pt}}m{#1}}

\newcommand*{\dif}{\mathop{}\!\mathrm{d}}

\newcommand{\E}{\mathbb{E}}

\newcommand{\namdpo}{\text{UDF-DPO}}
\newcommand{\namix}{\text{UDF-MIX}}

\renewcommand{\E}{\mathbb{E}}

\newcommand{\myparagraph}[1]{\noindent \textbf{#1}}

\theoremstyle{plain}
\newtheorem{theorem}{Theorem}[section]
\newtheorem{proposition}[theorem]{Proposition}

\theoremstyle{definition}

\theoremstyle{remark}

\title{Towards Universal Debiasing for Language Models-based Tabular Data Generation}

\author{
\textbf{Tianchun Li}\textsuperscript{1},
\textbf{Tianci Liu}\textsuperscript{1},
\textbf{Xingchen Wang}\textsuperscript{1},
\textbf{Rongzhe Wei}\textsuperscript{2},\\
\textbf{Pan Li}\textsuperscript{2},
\textbf{Lu Su}\textsuperscript{1},
\textbf{Jing Gao}\textsuperscript{1}\\[6pt]
\textsuperscript{1}Purdue University, West Lafayette, IN, USA\\
\textsuperscript{2}Georgia Institute of Technology, Atlanta, GA, USA\\[6pt]
\texttt{\{li2657, liu3351, wang2930, lusu, jinggao\}@purdue.edu}\\
\texttt{\{rongzhe.wei, panli\}@gatech.edu}
}

\begin{document}
\maketitle
\begin{abstract}
Large language models (LLMs) have achieved promising results in tabular data generation. However, inherent historical biases in tabular datasets often cause LLMs to exacerbate fairness issues, particularly when multiple advantaged and protected features are involved. In this work, we introduce a universal debiasing framework that minimizes group-level dependencies by simultaneously reducing the mutual information between advantaged and protected attributes. By leveraging the autoregressive structure and analytic sampling distributions of LLM-based tabular data generators, our approach efficiently computes mutual information, reducing the need for cumbersome numerical estimations. Building on this foundation, we propose two complementary methods: a direct preference optimization (DPO)-based strategy, namely \namdpo, that integrates seamlessly with existing models, and a targeted debiasing technique, namely \namix, that achieves debiasing without tuning the parameters of LLMs. Extensive experiments demonstrate that our framework effectively balances fairness and utility, offering a scalable and practical solution for debiasing in high-stakes applications.
\end{abstract}

\section{Introduction}
Large Language Models (LLMs)~\citep{lewis2019bart, brown2020language, kojima2022large, achiam2023gpt} demonstrate extraordinary ability to understand~\citep{jiang2020can}, reason~\citep{chang2024survey}, and generate text~\citep{ji2023survey}. These advancements have pushed new boundaries across a wide range of domains~\citep{yin2023survey, yang2024harnessing}. As one of the most common data forms~\citep{borisov2022deep}, there has been a growing trend to leverage LLMs for tabular data tasks, such as understanding~\cite{sui2024table}, prediction~\citep{ruan2024language}, and generation~\citep{borisov2023languagemodelsrealistictabular, zhao2023tabula, gulati2024tabmt}.

Despite their powerful capabilities, LLMs suffer from fairness issues when acting on tabular data, i.e., advantaged features (e.g., income) are often correlated with protected attributes (e.g., gender). Such biases widely exist in tabular data due to historical reasons~\citep{mehrabi2021survey}. Consequently, when LLMs are trained on this data, they will inherit existing biases~\citep{schick2021self}. Moreover, because the generated data is often used to train downstream prediction tasks for high-stakes domains such as job applications, the inherited bias raises fairness concerns for the downstream models as well~\citep{borisov2022deep}.

To address fairness concerns in LLMs, one approach is to adapt debiasing methods from non-LLM tabular data generators to ensure fairness in LLM-based generation. 
However, these existing methods only target bias between one advantage feature-protected attribute pair (e.g., \textit{income}-\textit{gender} pair) that adheres to only one downstream task~\citep{calmon2017optimized, xu2018fairganfairnessawaregenerativeadversarial, doi:10.1137/1.9781611976700.23, van2021decaf, abroshan2024imposing}.
That is to say, only the bias in \textit{gender} when predicting \textit{income} is guaranteed eliminated when using the generated dataset. 
When users want to work on other downstream tasks such as predicting \textit{education level}, fairness guarantee requires retraining the data generator again.

Yet, tabular datasets typically contain multiple advantaged features (e.g., income, education, occupation) and protected attributes (e.g., age, gender, race), making retraining for every possible pair computationally prohibitive. Another approach is to adapt for debiasing LLMs in text generation. Most existing methods focus on debiasing a single protected attribute~\citep{liu2021dexpertsdecodingtimecontrolledtext, yang2023unifieddetoxifyingdebiasinglanguage, liu2024lidao}. Therefore, these methods still cannot address settings with multiple protected attributes.

Rather than relying on pairwise debiasing methods, we propose a group-wise debiasing approach that eliminates all dependencies between advantaged features and protected attributes. Thus, our formulation partitions features into advantaged features (e.g., income, education, occupation), protected attributes (e.g., race, gender), and remaining features, and minimizes the group-level Mutual Information (MI) between the advantaged and protected features. Notably, pairwise debiasing is a special case of this broader framework, where the protected attribute and advantaged feature groups each contain only one feature. Additionally, the minimization of group-level MI also aligns with the principle of intersectional fairness~\cite{gohar2023survey} in the sense that, when mutual information is zero, every advantaged feature is guaranteed to be independent of any combination of protected features. While the minimization of MI provides fairness guarantees, breaking these dependencies inherently alters the learned distribution, potentially causing the generated data to deviate from the original. To prevent excessive distortion while still reducing bias, we impose an additional constraint that balances fairness against data utility during optimization. \textit{This universal debiasing framework for tabular data generation is our first key contribution.}

However, 
MI lacks a closed-form expression, making its computation challenging, let alone minimization for debiasing. 
This difficulty is exacerbated in high-dimensional spaces, where tabular data often lie in complex manifold~\citep{liu2024towards}. 
While this challenge cannot be solved in general, 
the unique auto-regressive nature of LLM-based tabular data generators allows us to derive efficient solutions for them. 
Specifically, LLMs generate different features of a tabular-typed sample one by one in a \textit{sequential} manner, and each feature is drawn from an \textit{analytic}-form distribution. 
Taking advantage of these analytic sampling distributions that are accessible, 
we propose a fine-tuning based solution for debiasing that eliminates the need for numerical estimation of MI. 
This solution can be readily implemented with direct preference optimization (DPO)~\citep{rafailov2024direct}, making our debiasing task no more difficult than standard fine-tuning. 
In addition, the debiased model maintains all applicability of the base LLM and can seamlessly replace the latter in all cases --- Notably, the fairness guarantee generalizes to diverse scenarios beyond data generation, such as data imputation. This strong \textit{one-for-all} guarantee makes our solution highly valuable. 
We refer to this DPO-based debiasing method as {{\namdpo}}. 

Built upon {\namdpo}, we derive {\namix}, a more efficient debiasing solution \textit{specialized for data generation}. 
{\namix} not only leverages the analytic sampling distribution, but also exploits the \textit{sequential} nature of the generation process. 
Specifically, {\namix} identifies \textit{a few} generation steps that cause the bias, and precisely alters these steps without changing others. 
This design leads to two remarkable efficiency improvements. 
First, as {\namix} only needs to debias a few generation steps, it relies on far fewer parameters, thereby achieving much better parameter efficiency. 
Second, through an innovative parametrization, 
we incorporate the fairness and utility balancing factor, which is usually treated as a hyper-parameter to tune, 
directly into {\namix} training. Consequently, {\namix} by design can handle the balance of conflicting fairness and utility without retraining, thereby substantially reducing the human burden and computation costs for tuning hyper-parameters for different tasks.
\textit{These two effective and efficient debiasing methods are also key contributions of our work.}

Our paper is organized as follows. Sec \ref{sec:prelim} and \ref{sec:limits} introduce preliminary and limitations of current methods. 
Sec \ref{sec:method} details our new universal debiasing framework and two effective solutions. 
Sec \ref{sec:experiment} presents extensive experiments to demonstrate the effectiveness of our methods. 
In the remaining part of this paper, we review related works in Sec \ref{sec:background}, and conclude the paper in Sec \ref{sec:conclusion}.


\section{Preliminary}\label{sec:prelim}
\myparagraph{Tabular Data.} Tabular data is structured in a table format, where each row corresponds to a sample and each column represents a feature, which can be of mixed types ~\citep{fang2024large,borisov2022deep}. 
Mathematically, a tabular dataset can be expressed as $D = \{d^{(i)} \}_{i=1}^N$, 
where each sample $d^{(i)}$ is a $K$-dimensional array. Each feature $d_k^{(i)}$ can be continuous, discrete, or unstructured, such as text descriptions\footnote{For brevity, the sample index $i$ will be ignored unless explicitly mentioned from now on. 
}.
Modeling tabular data is particularly challenging due to its heterogeneous feature types~\citep{sahakyan2021explainable, wang2024challenges, fang2024large}.
Traditional deep learning models are typically designed for a single data type, such as continuous-valued images or discrete textual data, and thus struggle to effectively handle tabular datasets~\citep{gorishniy2021revisiting, borisov2022deep, grinsztajn2022tree, chen2023inner}.

\myparagraph{Textual encoding of tabular data.} 
Recent works \citep{borisov2023languagemodelsrealistictabular, zhang2023generativetablepretrainingempowers} have demonstrated that the ability of LLMs to process diverse data types opens new avenues for modeling tabular data through the technique of \textit{textual encoding}. Specifically, given a feature $d_k$ with the name $f_k$, it can be represented as a short text in the form of ``{$f_k$ is $d_k$}'' (e.g., ``age is 20''). By concatenating all these texts into a single paragraph, a tabular dataset can be transformed into a textual representation, enabling standard LLMs to model it effectively. For simplicity, we refer to such text-encoded data as $D$.

\section{Bias in Tabular Data and Limitations of Pairwise Debiasing}\label{sec:limits}

Real-world tabular data often contains social bias from historical sources. 
For example, in credit application datasets, advantaged features such as \textit{income} and \textit{occupation} are often associated with \textit{genders}~\citep{caton2024fairness}. 
As a result, machine learning-based decision-makers trained on such biased datasets tend to discriminate \textit{female applicants} by predicting them as \textit{low income}, leading to \textit{fairness} concerns \citep{zemel2013learning,hardt2016equality,liu2023simfair}.
In response, existing works have proposed imposing some \textit{independence} 
between ML methods' action on the so-called \textit{advantaged} feature (\textit{income} in our example), and the demographic group \textit{gender} as a \textit{protected attribute} \citep{caton2024fairness}. Representative independence formulation (requirements) include Demographic Disparity (DP)~\citep{zemel2013learning} and Equalized Odds (EO)~\citep{hardt2016equality}.

Recent works showed that when generative models such as LLMs trained on biased datasets reproduce or even amplify such bias \citep{sui2024table}.
Consequently, 
when sharing such a data generator, the bias will be spread as well. 
This raises serious concerns for tabular data in high-stakes domains such as job applications, banking, and so on \citep{dastin2018amazon}. To prevent the bias in the generated data from propagating to downstream tasks, previous works impose fairness constraints when training the generative model. These constraints are specific to the advantaged feature (e.g., income) and protected attribute (e.g., gender) that will be used for downstream tasks.

However, if a downstream user is interested in a different pair of advantaged features and protected attributes (e.g., occupation and race) other than the ones used during training the generative model, 
the model must be retrained to address that new combination. Therefore, we refer to such methods as \textbf{Pairwise debiasing} to highlight that their fairness can only be guaranteed on a specific pair of advantaged features and protected attributes. However, the tabular data contains multiple advantaged features (e.g., income and occupation) and
protected attributes (e.g., race and gender). Such retraining for every possible pair of advantaged features and protected attributes is computationally infeasible for LLMs.

\section{Proposed Method}\label{sec:method}
\subsection{A Universal Debiasing Formulation}
Given that existing debiasing methods for tabular data generation are constrained by their specialized \textit{pairwise debiasing} design, it is necessary to employ a \textit{groupwise debiasing} approach in the sense that simultaneously debiases all advantaged features and protected attributes. In this light, we refer to such debiasing as universal debiasing. Our formulation starts with a key common sense based on the practical meaning of social bias: \textit{Given the interpretable nature of tabular datasets, the advantaged features and protected attributes are easy to identify}.

Based on this common sense, 
we split $K$ features $d_{1:K}$ into three groups.
First, $s$ is the collection of all protected features (e.g., gender and race). 
Second, $d_{as}$ is the collection of features that cannot be associated with $s$, and will raise fairness concerns otherwise (e.g., income level, education level, job eligibility).
Finally, $d_s$ denotes the remaining features that can freely vary across different $s$. 
Note that our categorization is a generalization of existing works, 
and reduces to the latter if $d_{as}$ and $s$ consist of only one feature respectively, where the single $d_{as}$ 
instantiates a \textit{label} to predict in a downstream task to be debiased.

Given tuple $(s, d_{as}, d_s)$, 
we define a group-level mutual information-based debiasing formulation. 
Suppose $p_\theta$ is a pre-trained data generator (such as an LLM), 
we quantify the bias carried by $p_\theta$ as
\begin{align*}
I_\theta (s, d_{as}) 
\triangleq
\E_{p_\theta} \left[ \log \frac{p_\theta(s, d_{as})}{p_\theta(s) p_\theta(d_{as})} \right],
\end{align*}
and propose to cast it into a \textit{fairer} generator $q_\phi$ by solving:
\begin{align}\label{eq:debias-obj}
\min\nolimits_{\phi} \ 
I_{\phi}(s, d_{as}) + \beta D_{KL}(p_\theta \|  q_\phi). 
\end{align}

Intuitively speaking, enforcing the first term (i.e., minimizing the MI) breaks the dependencies between two groups. Specifically, the benefits of using MI lie in two folds. 
First, 
mutual information as a bias measure is closely connected to the existing fairness notion demographic parity (DP)~\citep{zemel2013learning},
and implies the latter when $I_\phi(s, d_{as}) = 0$. 
Second, 
Eq \eqref{eq:debias-obj} extends debiasing from a single feature-level to 
a feature set-level, thereby imposing a stronger fairness guarantee for 
downstream applications. 
Specifically, 
any possible label $y \in d_{as}$ will be \textit{fair} with respect to every protected feature $a \in s$ thanks to the data processing inequality \citep{cover1999elements}
\begin{align*}
    I(s, d_{as}) \geq I(s, y) \geq I(a,y).
\end{align*}

The second term KL penalty restricts $q_\phi$ to stay close to base $p_\theta$, so that the data generated by $q_\phi$ has high quality \citep{kingma2013auto}.
Hyper-parameter $\beta$ balances the two terms and controls the \textit{fairness-utility trade-off}.

Eq \eqref{eq:debias-obj} provides a general debiasing framework that can be imposed on any data generators. However, this optimization is nontrivial to solve, due to the lack of a closed-form expression for mutual information that involves the high dimensional distribution $q_\phi$.

However, 
the auto-regressive nature of LLM allows one to freely control the feature generating orders. 
This flexibility offers us more effective ways to reduce the computational complexity of debiasing, as detailed below.

\subsection{Debiasing Through Finetuning}
As mentioned above, the special generating process from LLMs enables effective debiasing. This section details a finetuning-based formulation and its solution.

Specifically, we reformulate the bias as a (negative) reward that the LLM should minimize, and cast debiasing from Eq \eqref{eq:debias-obj} into a preference optimization problem, so that direct preference optimization (DPO) and its variants with different parameter-efficient fine-tuning strategies can be applied \citep{ethayarajh2024kto, azar2024general, guo2024controllable, hu2021loralowrankadaptationlarge, wang2024roselorarowcolumnwisesparse, zhong2025neatnonlinearparameterefficientadaptation, chen2024parameter, liu2025unlocking}. 
Mathematically speaking, we have
\begin{align}\label{eq:reward}
I_{\phi}(s, d_{as}) 
&= 
\mathbb{E}_{{q_\phi}} \left[ \log\frac{q_\phi(s, d_{as})}{q_\phi(s)q_\phi(d_{as})}\right] \notag \\
&= 
\mathbb{E}_{{q_\phi}} \left[ \log\frac{q_\phi(d_{as} \mid s)}{q_\phi(d_{as})}\right] \notag \\
&\triangleq \mathbb{E}_{{q_\phi}} [- r(s, d_{as})]. 
\end{align}
Here the negative reward $-r(s, d_{as})$ measures 
\textit{to what extent knowing protected features $s$ helps predict $d_{as}$}.
A high reward indicates that $s$ and $d_{as}$ are essentially independent, thus the generated data are fair. 
Built upon this, Eq \eqref{eq:debias-obj} can be written as a standard preference optimization objective with forward KL \footnote{Note that we flip the minimization to maximization.}
\begin{align}\label{eq:dpo-obj}
    \max\nolimits_\phi \mathbb{E}_{{q_\phi}} [r(s, d_{as})] - \beta D_{KL} (p_\theta \| q_\phi),
\end{align}
This objective can be optimized in either an on-policy or off-policy way, and we conduct an \textit{approximately} on-policy learning with DPO. 
Specifically, after several DPO fine-tuning steps, 
we recollect a new dataset from current $q_\phi$. Next, we compute a reward for each sample based on Eq \eqref{eq:reward}. 
Finally, we randomly construct pairs of samples whose reward gap exceeds a pre-specified threshold. The sample that achieves a \textit{higher} reward is treated as the \textit{preferred} one. 
The next round of DPO fine-tuning is conducted on the new dataset. 
We dub our method \textit{\namdpo}. The complete algorithm is summarized in Algorithm \ref{alg:udf-dpo}.

We conclude this section with two remarks. 
First, $d_{as}$ and $s$ are symmetric in $I_\phi(s, d_{as})$; therefore, one can also define the reward as the log ratio between $q_\phi(s \mid d_{as})$ and $q_\phi(s)$ without violating the validity of our framework. 
Second, the key flexibility that auto-regressive LLMs offers is that we can directly compute all required probabilities (and the reward) analytically. In contrast, for other generators, these quantities have to be estimated numerically.

\subsection{Adaptive-Inference Time Debiasing}
Computing Eq. \eqref{eq:dpo-obj} analytically offers an additional benefit: it preserves the flexibility of the LLM by maintaining the free control of feature generating orders. However, this flexibility is mostly beneficial to tasks beyond generation tasks such as data imputation.

In this section, we show that by sacrificing some of this flexibility, we can further reduce the computational complexity in two means. First, we can further reduce the complexity in computing the debiasing object by focusing on an intermediate part of the generation process. Second, we can enhance the LLM’s generation process with a lightweight module that adapts to different hyperparameter settings for $\beta$ without requiring retraining, thus achieving inference-time debiasing.

Specifically, an autoregressive LLM allows us to \textit{generate data} according to the decomposed order\footnote{We abuse the notation a bit by expressing different distributions as the function of the same parameters. }
\begin{align*}
    p_\theta(s, d_{as}, d_s) = p_\theta (s) p_\theta (d_{as} \mid s) p_\theta (d_s \mid s, d_{as}). 
\end{align*}
Note that only the second term $p_\theta (d_{as} \mid s)$ affects the fairness, and $d_s$ by definition can be generated freely. 
Therefore, instead of altering the complete generating process of LLM $p_\theta$, 
we solve Eq \eqref{eq:debias-obj} by only replacing the intermediate $p_\theta(d_{as} \mid s)$ with one that minimizes the debiasing objective.
This leads to 
\begin{align}\label{eq:dec-obj}
\min\nolimits_\phi 
&\quad I_\phi (s, d_{as}) + \beta D_{KL} ( p_\theta \| q_\phi ) \notag \\
\text{s.t.} &\quad 
q_\phi(s, d_{as}, d_{s}) \triangleq p_\theta(s) \times \notag \\ 
&\quad\quad 
\underbrace{q_\phi (d_{as} \mid s)}_{\text{learnable}} 
p_\theta(d_s \mid s, d_{as}). 
\end{align}
Training a $q_\phi(d_{as} \mid s)$ from scratch can be expensive especially when $d_{as}$ and $s$ are of high dimensions. 
To avoid this computational burden, 
we propose a reparameterized form based on the following proposition, with its proof deferred to App \ref{app:proof}.

\begin{proposition}\label{prop:optimal-dis}
Consider the optimization problem given in Eq \eqref{eq:dec-mix}.
Then $p_\theta(d_{as})$ and $p_\theta(d_{as} \mid s)$ achieve the optimal utility under strict or no fairness constraints, respectively. 
Specifically, we have 
\begin{align*}
p_\theta(d_{as}) 
=& 
\arg \min\nolimits_{q_\phi(d_{as} \mid s)}  \left\{D_{KL}[p_\theta \| q_\phi ] \right\} \\
&
\text{s.t.}
\quad
I_\phi(s, d_{as}) = 0,
\end{align*}
and 
\begin{align*}
p_\theta(d_{as} \mid s) = \arg \min\nolimits_{q_\phi}  D_{KL}[p_\theta \| q_\phi].
\end{align*}
\end{proposition}

Given the optimal solutions from Prop~\ref{prop:optimal-dis}, 
it is viable to strike a balance between fairness and utility at efficiency by combining  them linearly~\citep{chuang2021fair,zhou2024counterfactual}.
To this end, 
we parameterize $q_\phi$ in Eq \eqref{eq:dec-obj} as a convex combination of them
\begin{align}\label{eq:dec-mix}
q_\phi (d_{as} \mid s) 
&= 
\lambda (s, \beta) p_\theta(d_{as}) + \notag \\
&\quad (1 - \lambda (s, \beta)) p_\theta(d_{as} \mid s),
\end{align}
and \textit{learn} the mixing weight $\lambda(s, \beta) \in [0, 1]$ only, which is a function of both $s$ and $\beta$.
Notably, its dependency on $s$ allows different level of debiasing strength for each different values of the protected attribute $s$. The larger values of $\lambda$ will be assigned to groups exhibiting stronger bias and vice versa. Such a targeted mixing strategy allows a fine-grained control over the fairness and utility tradeoff. At the same time, $\lambda$ as a function of hyper-parameter $\beta$ essentially enhances overall computation efficiency by avoiding multiple rounds of retraining when adjusting $\beta$.
In practice, we parameterize $\lambda(\cdot, \cdot)$ with a lightweight MLP. The objective is again trained with DPO loss as presented before.
The complete algorithm is summarized in Algorithm ~\ref{alg:adaptive-debiasing}. 

While the fairness-utility trade-off is widely observed in general, 
our mixing-typed solution strikes an effective balance as revealed by the following theorem. See its proof in Appendix \ref{app:proof}.
\begin{theorem}\label{thm:total-drop}
When using Eq \eqref{eq:dec-mix}, the fairness-utility total loss is upper bounded.
Specifically
\begin{align*}
    I_\phi(s, d_{as}) + D_{KL}(p_\theta \| q_\phi) 
    \leq I_{\theta}(d_{as}, s).
\end{align*}
\end{theorem}

Notably, Thm \ref{thm:total-drop} shows that 
while increasing fairness may lead to a drop in utility and vice versa, this trade-off is \textit{efficient} in the sense that their total degradation is bounded.

\section{Experiments}
\label{sec:experiment}
We evaluate our methods on two practical use cases with generated data from three diverse tabular datasets, each featuring different protected attributes and advantaged features. Our methods achieve debiasing between multiple potential target variables and protected attributes while preserving high data utility. 


\subsection{Experiment Setup}
\myparagraph{Backbone Tabular Data Generator.}
We use GReaT~\citep{borisov2023languagemodelsrealistictabular} as the backbone tabular generator. We follow the choice of using GPT-2~\citep{radford2019language} as the base LLM. 

\myparagraph{Baselines and Implementation Details. } For tabular data generation, we compare our debiasing methods with four baselines: \textbf{GReaT} (the backbone generator), \textbf{DECAF-DP}, a variant of DECAF~\citep{van2021decaf} focusing on demographic disparity, and two GAN-based generators, \textbf{TabFairGAN}~\citep{rajabi2022tabfairgan} and \textbf{FairGAN}~\citep{xu2018fairganfairnessawaregenerativeadversarial}. We refer to the downstream model trained on real data as ``Original''. All the experiments are run on RTX A6000 GPUs. More details are given in the Appendix.

\myparagraph{Datasets. } We evaluate our model using three diverse datasets. The Adult dataset \cite{becker1996adult} contains 11 attributes. We choose \textit{race} and \textit{gender} as potential protected attributes $s$, and \textit{income} and \textit{education} as $d_{as}$. The Credit Approval dataset \cite{quinlan1987credit} contains 15 features. The potential protected attributes $s$ include \textit{gender} and \textit{race}. For potential target variables, we include \textit{approval} and \textit{employment status} as $d_{as}$. The Student Performance dataset contains 30 attributes. We choose the $s$ as \textit{Mother's Job (MJ)}, \textit{Father's Job (FJ)}, \textit{Age}, and \textit{Gender}. The $d_{as}$ are \textit{First Period (1\textsuperscript{st}) Grade} and \textit{Second Period (2\textsuperscript{nd}) Grade}. 

\myparagraph{Scalability of Datset Size. }Notably, our debiasing methods are not constrained by the size of the datasets. Instead, the scalability lies in the pre-trained models in the sense that our framework is trained using the generated data, not the real data. 

\myparagraph{Tabular Tasks and Evaluations. } Based on the practical usage of the generated data, we consider the two tasks, evaluated from two dimensions: fairness and data utility. We further evaluate the efficiency of our methods. 
\begin{itemize}
\item \textbf{Tabular Data Generation for Predictive Downstream Tasks:} We establish the downstream task by pairing chosen variables from $d_{as}$ as target variables and $s$ as protected attributes. For each of these pairings, we train a MLP as the prediction model on the generated dataset to predict the target variables and evaluate data utility via accuracy (\textbf{Acc.}) and \textbf{AUROC}. Fairness is evaluated in three ways: estimated Mutual Information (\textbf{MI}) between the protected attributes and the model’s prediction on target variables.
Demographic Parity (\textbf{DP}), quantified as the total variation distance between prediction distributions across groups~\citep{van2021decaf}; and Equalized Odds (\textbf{EO}), calculated by the maximum disparity in true positive and false positive rates among all groups~\citep{hardt2016equality}.

\item \textbf{Tabular Data Missing Value Imputation:} Since the LLM-based generator can generate features based on observed features, it is used for filling missing values in the tabular dataset. We follow the Missing Completely At Random  (MCAR)~\citep{little1988test} setting, where each feature has a certain probability of being marked as missing. We set the missing probability to $0.4$. For fairness, we estimate the \textbf{MI} between $d_{as}$ and $s$ in the generated data. For data utility, we measure averaged \textbf{RMSE} over all missing continuous features and averaged Accuracy over all categorical features. However, in some rows, $d_{as}$ and $s$ might not be marked as missing, which means the bias already exists and cannot be reduced.

\item \textbf{Efficiency:} We further evaluate the Efficiency of our methods by the time of measuring training and generation (in seconds) for different generation sizes. 

\end{itemize}

\begin{table*}[htb!]
\centering
\resizebox{\linewidth}{!}{%
\begin{tabular}{r ccccc c ccccc}
\toprule[0.3ex]
& \multicolumn{2}{c}{\textbf{Utility $\uparrow$}} & \multicolumn{3}{c}{\textbf{Bias $\downarrow$}} && \multicolumn{2}{c}{\textbf{Utility $\uparrow$}} & \multicolumn{3}{c}{\textbf{Bias $\downarrow$}} \\
\cmidrule(lr){2-3} \cmidrule(lr){4-6} \cmidrule(lr){8-9} \cmidrule(lr){10-12}
& \textbf{Acc.} & \textbf{AUROC} & \textbf{MI} & \textbf{DP} & \textbf{EO} && \textbf{Acc.} & \textbf{AUROC} & \textbf{MI} & \textbf{DP} & \textbf{EO} \\
\midrule[0.2ex]
\multicolumn{12}{l}{\qquad \qquad \qquad Task 1: Gender-Income \qquad \qquad \qquad\qquad \qquad \qquad \qquad \quad Task 2: Race-Education Level}\\
\midrule[0.2ex]
Real Data   & 84.12$_{0.22}$          & 90.46$_{0.71}$          & 2.52$_{0.19}$          & 19.78$_{1.71}$         & 11.17$_{0.59} $        && 69.79$_{1.21}$       & 76.87$_{1.26}$         & 0.93$_{0.29}$         & 7.31$_{0.39}$          & 6.17$_{0.61}$          \\
GReaT        & 84.32$_{0.15}$ & 89.37$_{0.30}$ & 7.01$_{0.12}$ & 17.29$_{1.83}$ & 19.76$_{3.44}$ && 67.63$_{0.04}$ & 74.14$_{0.09}$ & 0.60$_{0.08}$ & 7.12$_{0.68}$ & 9.03$_{0.71}$ \\
\midrule[0.2ex]
DECAF-DP    & 75.95$_{0.10}$ & 86.79$_{0.32}$ & 0.04$_{1.42}$ & \textbf{1.12$_{0.23}$} & \textbf{2.40$_{0.51}$} && 57.47$_{0.55}$ & 58.50$_{1.08}$ & 0.80$_{0.91}$ & 9.34$_{1.90}$ & 10.93$_{1.94}$ \\
TabFairGAN  & 80.59$_{0.30}$ & 83.44$_{0.26}$ & \textbf{0.01$_{0.01}$} & 4.22$_{1.03}$ & 19.28$_{1.56}$ && \textbf{68.40$_{0.23}$} & \textbf{75.03$_{0.20}$} & 1.60$_{0.07}$ & 8.14$_{0.91}$ & 7.57$_{1.21}$ \\
FairGAN     & 75.70$_{1.77}$ & 74.37$_{1.89}$ & \underline{0.02$_{0.01}$} & 6.28$_{3.02}$ & 10.27$_{7.59}$ && 44.39$_{0.85}$ & 48.34$_{3.57}$ & 1.12$_{0.32}$ & 22.91$_{3.92}$ & 25.02$_{4.56}$ \\

\rowcolor{gray!10}{\namdpo} & & & & & && & & & & \\
$\beta=0.1$ & 76.44$_{0.21}$ & 81.69$_{0.38}$ & 0.30$_{0.03}$ & \underline{1.39$_{0.28}$} & \underline{2.64$_{0.87}$} && 66.34$_{0.14}$ & 68.19$_{0.42}$ & \textbf{0.29$_{0.11}$} & \textbf{1.97$_{0.31}$} & \textbf{3.14$_{0.49}$} \\
$\beta=1$   & 81.71$_{0.38}$ & 86.04$_{0.43}$ & 1.20$_{0.03}$ & 9.02$_{1.96}$ & 5.73$_{2.13}$ && 65.33$_{0.53}$ & 71.82$_{0.62}$ & 0.43$_{0.06}$ & 5.38$_{2.63}$ & 6.27$_{2.42}$ \\
$\beta=10$  & \underline{82.01$_{0.30}$} & \textbf{87.01$_{0.19}$} & 1.45$_{0.07}$ & 9.21$_{1.03}$ & 5.78$_{0.97}$ && 66.43$_{0.75}$ & \underline{73.83$_{1.72}$} & 0.54$_{0.07}$ & 8.25$_{0.56}$ & 8.33$_{0.64}$ \\

\rowcolor{gray!10}{\namix} & & & & & && & & & & \\
$\beta=0.1$  & \textbf{82.08$_{0.23}$} & 86.39$_{0.37}$ & 0.02$_{0.02}$ & 5.99$_{1.22}$ & 11.84$_{4.94}$ && 66.29$_{0.46}$ & 72.29$_{0.35}$ & 0.37$_{0.02}$ & 3.35$_{1.70}$ & 4.46$_{0.93}$ \\
$\beta=1$    & 81.96$_{0.41}$ & 86.35$_{0.17}$ & 0.10$_{0.03}$ & 5.54$_{1.08}$ & 10.90$_{2.54}$ && 65.67$_{0.29}$ & 72.10$_{0.19}$ & 0.38$_{0.01}$ & 7.99$_{1.44}$ & 7.49$_{1.39}$ \\
$\beta=10$   & \underline{81.94$_{0.47}$} & \underline{86.95$_{0.31}$} & 0.29$_{0.09}$ & 7.48$_{2.53}$ & 7.56$_{2.32}$ && \underline{66.63$_{0.24}$} & 72.31$_{0.16}$ & 0.40$_{0.04}$ & 3.47$_{2.51}$ & 6.04$_{1.83}$ \\
\midrule[0.3ex]

\multicolumn{12}{l}{\qquad \qquad \qquad Task 1: Age, MJ, FJ, Gender – 1\textsuperscript{st} Grade \qquad\qquad \quad \quad \quad \quad Task 2: Age, MJ, FJ, Gender – 2\textsuperscript{nd} Grade}\\
\midrule[0.2ex]

Real Data   & 87.32$_{0.29}$          & 90.27$_{0.14}$           & 6.24$_{0.04} $         & 8.93$_{1.21}$          & 9.14$_{0.29}$          && 96.14$_{0.61}$        & 98.32$_{0.16}$         & 8.41$_{0.38}$         & 9.43$_{0.15}$          & 9.02$_{0.61}$          \\
GReaT        & 85.23$_{3.87}$ & 88.47$_{2.48}$ & 5.41$_{1.22}$ & 7.02$_{1.24}$ & 8.19$_{1.42}$ && 94.31$_{3.51}$ & 96.72$_{1.48}$ & 4.51$_{1.41}$ & 8.24$_{2.04}$ & 9.41$_{1.71}$ \\

\midrule[0.2ex]
DECAF-DP     & 62.19$_{6.29}$ & 65.92$_{4.22}$ & \textbf{0.98}$_{0.01}$ & \textbf{3.01}$_{1.28}$ & \textbf{2.21}$_{1.28}$ && 72.19$_{4.12}$ & 78.41$_{1.48}$ & \textbf{2.01}$_{0.47}$ & \underline{5.81$_{1.25}$} & \textbf{6.14}$_{1.48}$ \\
TabFairGAN   & 78.42$_{2.19}$ & 82.45$_{3.19}$ & 3.81$_{1.91}$ & 5.89$_{1.29}$ & 6.79$_{1.92}$ && 88.32$_{3.12}$ & 92.41$_{1.33}$ & 6.29$_{1.49}$ & 8.41$_{1.29}$ & 9.12$_{1.44}$ \\
FairGAN      & 76.42$_{3.18}$ & 84.23$_{3.29}$ & 2.48$_{1.29}$ & 6.28$_{1.29}$ & 7.29$_{1.93}$ && 89.19$_{4.12}$ & 90.33$_{1.64}$ & 5.71$_{1.29}$ & 8.79$_{1.81}$ & 9.42$_{1.73}$ \\

\rowcolor{gray!10}{\namdpo}  & & & & & && & & & &\\
$\beta=0.1$ & \textbf{82.70}$_{3.16}$ & \textbf{87.46}$_{2.59}$ & 2.39$_{0.27}$ & 6.51$_{0.92}$ & 7.14$_{0.92}$ && 90.41$_{2.14}$ & 94.03$_{2.49}$ & 2.26$_{0.61}$ & 6.11$_{1.05}$ & 6.78$_{1.26}$ \\
$\beta=1$   & 78.82$_{1.57}$ & 85.91$_{1.53}$ & 1.27$_{0.39}$ & \underline{5.07$_{1.71}$} & 6.49$_{1.12}$ && 90.21$_{1.31}$ & 94.72$_{1.82}$ & 2.12$_{0.49}$ & 6.21$_{1.51}$ & 7.41$_{1.25}$ \\
$\beta=10$  & 78.36$_{1.97}$ & 85.98$_{2.24}$ & 1.39$_{0.56}$ & 5.28$_{1.29}$ & 6.83$_{2.84}$ && \underline{90.95$_{1.79}$} & \underline{95.20$_{0.64}$} & 2.26$_{0.84}$ & 7.53$_{1.72}$ & 7.91$_{1.72}$ \\

\rowcolor{gray!10} {\namix} & & & & & && & & & & \\
$\beta=0.1$ & 78.31$_{2.19}$ & 85.13$_{3.27}$ & \underline{1.12$_{0.18}$} & 5.26$_{0.28}$ & 6.41$_{0.29}$ && 89.01$_{3.27}$ & 94.31$_{2.81}$ & \underline{2.04$_{0.71}$} & \textbf{5.73}$_{0.81}$ & \underline{6.21$_{1.14}$} \\
$\beta=1$   & 79.38$_{2.46}$ & 86.02$_{4.81}$ & 1.53$_{0.23}$ & 6.19$_{1.02}$ & 6.82$_{1.27}$ && 90.32$_{2.81}$ & 94.47$_{1.92}$ & 2.42$_{0.53}$ & 6.14$_{1.26}$ & 7.31$_{1.69}$ \\
$\beta=10$  & \underline{79.52$_{2.18}$} & \underline{86.82$_{2.61}$} & 1.62$_{0.42}$ & 5.97$_{1.16}$ & \underline{6.09}$_{1.92}$ && \textbf{91.21}$_{1.28}$ & \textbf{95.21$_{1.27}$} & 3.04$_{0.13}$ & 7.32$_{1.84}$ & 7.71$_{1.24}$ \\

\bottomrule[0.3ex]
\end{tabular}
}
\caption{
Performance on the Adult (upperhalf) and Student Performance (lowerhalf) datasets for two downstream tasks each.
Only Task 1's target and protected features are revealed to task-specific baselines during training, whereas our methods debias all potential downstream tasks simultaneously.
Best results are in \textbf{bold} and second-best results are \underline{underlined}.
\textbf{Baseline methods trained to debias Task 1 remain unfair on Task 2.}
}
\vspace{-4mm}
\label{tab:combined-side-by-side}
\end{table*}

\subsection{Performance on Tabular Data Generation for Predictive Downstream Tasks. }
Table~\ref{tab:combined-side-by-side} presents results for the Adult and Student Performance datasets. For the Adult dataset, Task 1 predicts whether income exceeds $50K$ with gender as the protected attribute, while Task 2 predicts whether education level exceeds high school with race as the protected attribute. For the Student Performance dataset, we use the same protected attributes, Age, MJ, FJ, and Gender—for both tasks; Task 1 targets first‐period grade, and Task 2 targets second‐period grade. Additional results on the Credit Approval are provided in the appendix. Only Task 1 is revealed for training task-specific baselines, whereas our methods can simultaneously debias across all possible downstream tasks.

\myparagraph{Debiasing and Utility trade-off. } In all downstream tasks in Table~\ref{tab:combined-side-by-side}, our methods achieve bias reduction while maintaining high data utility when compared to GReaT. Specifically, for $\namix$ debiasing method, when $\beta = 0.1$, it reduces the bias significantly compared with GReaT while maintaining similar predictive performance. For $\namdpo$ debiasing, similar phenomenon is achieved when $\beta=1$. However, when compared with task-specific debiasing methods, task-specific baselines like  DECAF-DP achieve generally satisfying data utility compared with similar debiasing scores. This is because the task-specific baselines are given the specific information that the downstream task will predict; for example, income and the corresponding protected attribute is gender. However, the task-specific baselines cannot guarantee fairness performance when the generated data is used for other prediction tasks by observing the performance decay in task 1 and task 2. We further perform a detailed trade-off analysis in the apendix. 

\myparagraph{Universal Debiasing performance. } 
By comparing Task 1 and Task 2 in Table~\ref{tab:combined-side-by-side}, our methods demonstrate universal debiasing across multiple downstream tasks. Specifically, when $\beta=0.1$, the $\namix$ method achieves significant bias reduction on both tasks, and $\namdpo$ attains similar performance when $\beta=1$. In contrast, task-specific benchmarks fail to guarantee fairness or data utility when applied to different downstream tasks. For example, DECAF-DP—despite achieving the best DP score in Task 1—performs poorly in Task 2, because it focuses solely on bias between income and gender in the Adult dataset and does not eliminate bias between education level and race.


\myparagraph{Bias in the original dataset. } As shown in Table~\ref{tab:combined-side-by-side}, when the downstream model is trained on the original dataset, it often produces the most biased yet most accurate predictions. Specifically, the model trained on real data attains the highest DP score, indicating greater bias than all other benchmarks, while also achieving the highest AUROC. 

\myparagraph{Bias in the LLM-based tabular generator. } Both sections of Table~\ref{tab:combined-side-by-side} show that data generated by GReaT exhibits similar or greater bias than the real data. Specifically, the estimated MI in GReaT-generated data is nearly three times higher than in the real data (Task 1). This likely explains why the EO of the downstream model trained on GReaT data exceeds the EO of the model trained on real data.


\begin{table}[htb!]
\definecolor{verylightgray}{gray}{0.9}

\centering
\resizebox{0.8\linewidth}{!}{
\begin{tabular}{r cc c}
\toprule[0.3ex]
& \multicolumn{2}{c}{\textbf{Utility $\uparrow$}} & \textbf{Bias $\downarrow$} \\ 
\cmidrule(lr){2-3} \cmidrule{4-4}

& \textbf{Acc.} & \textbf{RMSE} & \textbf{MI} \\
\cmidrule(lr){2-3} \cmidrule{4-4}
Original        & $-$                           & $-$                           & 23.91 \\

\noalign{\vskip 0.2ex}\cdashline{2-4}\noalign{\vskip 0.2ex}
GReaT           & 60.08 $\pm$ 0.42           & 15.12 $\pm$ 0.08           & 18.56 $\pm$ 0.40 \\

\noalign{\vskip 0.2ex}\cdashline{2-4}\noalign{\vskip 0.2ex}

\rowcolor{gray!10} {$\namdpo$} & & & \\
$\beta = 0.1$   & 56.45 $\pm$ 0.28          & 16.67 $\pm$ 0.13           & 15.44 $\pm$ 0.61 \\
$\beta = 1$     & 62.63 $\pm$ 0.60          & 16.41 $\pm$ 0.07           & 15.31 $\pm$ 1.01 \\
$\beta = 10$    & 61.50 $\pm$ 0.32          & 16.94 $\pm$ 0.22           & 15.30 $\pm$ 0.70 \\


\rowcolor{gray!10} {$\namix$} & & & \\
$\beta = 0.1$   & 47.44 $\pm$ 0.22           & 39.87 $\pm$ 41.29          & 15.38 $\pm$ 0.70 \\
$\beta = 1$     & 47.28 $\pm$ 0.65           & 15.91 $\pm$ 0.09           & 14.89 $\pm$ 0.64 \\
$\beta = 10$    & 47.68 $\pm$ 0.16           & 16.08 $\pm$ 0.13           & 15.33 $\pm$ 0.58 \\
\bottomrule[0.3ex]

\end{tabular}
}
\caption{
Data imputation performance on Adult dataset.
}
\label{tab:ad-di}
\vspace{-0.2cm}
\end{table}

\subsection{Data Imputation}
In the data imputation task, we impute missing values five times with different random seeds and report the mean and standard deviation in Table~\ref{tab:ad-di}. Table~\ref{tab:ad-di} shows that, under a similar $\beta$, our debiasing methods outperform GReaT at comparable fairness levels. Specifically, estimated MI, a measure of dataset bias, is lower for both $\namdpo$ and $\namix$ than for GReaT, indicating they maintain debiasing when filling in missing values.

\subsection{Overall Performance Comparison between $\namix$ and $\namdpo$}
One possible reason that the $\namdpo$ generally performs better than $\namix$ is that $\namdpo$ is more flexible during debiasing than $\namix$, as it offers more modification options during the generation process. According to Eq.~\ref{eq:reward}, $\namdpo$ can reduce the mutual information by modifying $q_\phi(d_{as}\mid s)$, $q_\phi(d_{as})$, or $q_\phi(s)$. Modifying the latter two introduces fewer disruptions to the correlations between features, while still lowering the mutual information. In contrast, $\namix$ is designed to modify only the intermediate $q_\phi(d_{as}\mid s)$. 

\subsection{Efficiency}
\begin{figure}[ht]
    \centering
    \includegraphics[width=\linewidth]{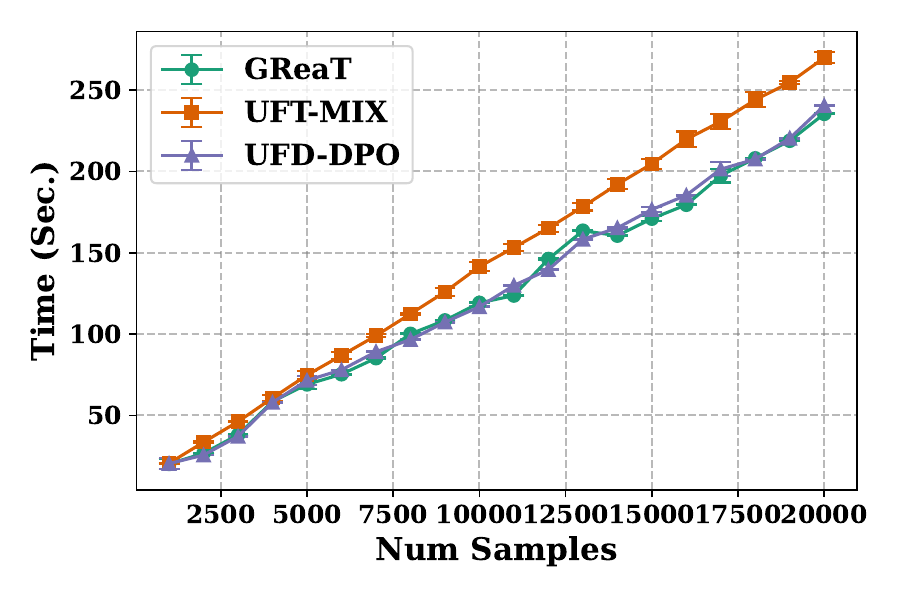}
    \vspace{-9mm}  
    \caption{
    Running time of base and debiased models. 
    Our methods add marginal computation overhead to data generation. 
    }
    \label{fig:generation_time}
    \vspace{-1mm}
\end{figure}

We measure both training and generation efficiency (in seconds) for each method in Table~\ref{tab:run-time} and Figure~\ref{fig:generation_time}. Since $\beta$ in $\namdpo$ does not affect efficiency, we fix $\beta=1$ and train for five epochs—its typical convergence point. For $\namix$, we sample 1,000 $\beta$ values to train only the lightweight MLP adapter, yielding faster training (Table~\ref{tab:run-time}). In generation, however, $\namix$ is slightly slower than $\namdpo$ and GReaT due to its extra layer of randomness (Figure~\ref{fig:generation_time}). $\namdpo$ and GReaT share the same generation process and thus exhibit similar generation efficiency.

\begin{table}[h]
\small
\begin{tabular}{lcc}
\toprule
& {$\namdpo$} & {$\namix$} \\
\midrule
Time & $399.56 \pm 3.85$ & $65.32 \pm 1.72$ \\
\bottomrule
\end{tabular}
\vspace{-1mm}
\centering
\caption{Finetuning time (s) of our methods.}\label{tab:run-time}
\end{table}
\vspace{-4mm}

\section{Related Work}\label{sec:background}
\myparagraph{LLM-based Tabular data Generation. } Besides GReaT~\citep{borisov2023languagemodelsrealistictabular}, \citet{zhao2023tabula} further shortens the textual encoding in the GReaT. \citet{zhang2023generativetablepretrainingempowers} fine-tunes the LLM from tabular data generation to classification. Alternatively, \citet{wang2024harmonic} combines the tabular data with clustering algorithms. However, all these LLM-based tabular data generators share the same fairness concern when generating tabular data. 

\myparagraph{Debiasing for Tabular data Generation. }
Generative Adversarial Networks (GAN)~\citep{goodfellow2020generative} are a popular choice for generating fair tabular data. \citet{xu2018fairganfairnessawaregenerativeadversarial} propose that, after training the GAN for tabular data generation, the generator can be further trained for fairness. \citet{rajabi2022tabfairgan, abroshan2024imposing} further utilize the discriminator to add the fairness constraint. \citet{van2021decaf} propose an inference time debiasing method. However, all these methods are formulated and designed to debias for specific protected attributes and target variables. 

\myparagraph{Debiasing for Text Generation in LLMs. }
Decoding-time debiasing approaches, as proposed by \citet{liu2021dexperts, yang2023unifieddetoxifyingdebiasinglanguage}, are closely related to tabular data generation. ~\citet{liu2024lidao} proposes a debiasing method that targets balancing the trade-off between fluency and bias mitigation. ~\citet{li2023guiding} uses a prompt-based method to guide the LLMs.

\section{Conclusion}\label{sec:conclusion}
We propose a universal debiasing framework for LLM-based tabular data that balances the fairness-utility trade-off for multiple advantaged features and protected attributes. Our DPO-based method, {\namdpo}, and the efficient adaptive approach, {\namix}, mitigate bias while preserving high data quality. Mathematical insights and experiments confirm that our approach outperforms existing pairwise methods, offering robust and scalable debiasing for high-stakes applications.

\section*{Limitations}
Our method UDF-Mix has additional computational overhead by requiring multiple $\beta$ values to be sampled and fit. However, our experiments show that restricting $\beta$ to the range [0, 50] is sufficient to achieve universal debiasing, which helps mitigate the impact of this overhead. Another limitation is that, for each dataset and each tabular data generator, our methods need to be re-trained. One future direction is achieving the debiasing with training across multiple datasets.

\section*{Acknowledgment}
This work is supported in part by the US National Science
Foundation under grant NSF IIS-2141037. Any opinions,
findings, and conclusions or recommendations expressed in
this material are those of the author(s) and do not necessarily
reflect the views of the National Science Foundation.

\bibliography{writing/reference}


\onecolumn
\appendix
\appendix

\section{Omitted Proof}\label{app:proof}

In this section we present the proof of theorems omitted in the main body.

\begin{proposition}
Consider the optimization problem given in Eq \eqref{eq:dec-mix}.
$p_\theta(d_{as})$ achieves the optimal fairness, and $p_\theta(d_{as} \mid s)$ achieves the optimal utility. 
Specifically, we have 
\begin{align}\label{eq:opt-fair}
p_\theta(d_{as}) = \arg \min\nolimits_{q_\phi(d_{as} \mid s)}  \left\{D_{KL}[p_\theta (s, d_{as}, d_s) \| q_\phi (s, d_{as}, d_s) ] : I_\phi(s, d_{as}) = 0 \right\} ,
\end{align}
and 
\begin{align}\label{eq:opt-util}
p_\theta(d_{as} \mid s) = \arg \min\nolimits_{q_\phi}  D_{KL}[p_\theta (s, d_{as}, d_s) \| q_\phi (s, d_{as}, d_s)].
\end{align}
\end{proposition}

\begin{proof}
Eq \eqref{eq:opt-util} can be verified directly by definition. 
To prove Eq \eqref{eq:opt-fair}, first note that when 
\begin{align*}
q_\phi(s, d_{as})
&= 
\int q(s, d_{as}, d_s) \dif d_s \\
&= 
\int p_\theta(s) p_\theta(d_{as}) p_\theta(d_s \mid s, d_{as}) \dif {d_s} \\
&= 
p_\theta(s) p_\theta(d_{as}) \int p_\theta(d_s \mid s, d_{as}) \dif {d_s} \\
&=
p_\theta(s) p_\theta(d_{as}).
\end{align*}
Namely, we have that $s$ and $d_{as}$ are independent, therefore, $I_\phi(s, d_{as}) = 0$. 
In addition, for any \textit{fair} $q_\phi$, 
\begin{align*}
D_{KL}(p_\theta \|  q_\phi) 
&=\mathbb{E}_{p_\theta} \left[ \log\left( \frac{p_\theta(s) p_\theta(d_{as} \mid  s) p_\theta(d_s  \mid  s, d_{as})}{p_\theta(s)q_\phi(d_{as} \mid s)p_\theta(d_s\mid s, d_{as})} \right) \right]\\
&= 
\mathbb{E}_{p_\theta}\left[\log\left( \frac{p_\theta(s)}{p_\theta(s)}\right)+ \log\left(\frac{p_\theta(d_{as}\mid s)}{q_\phi(d_{as}\mid s)} \right) + \log\left( \frac{p_\theta(d_s  \mid  s, d_{as})}{p_\theta(d_s  \mid  s, d_{as})}\right)   \right]\\
&=
\mathbb{E}_{p_\theta}\left[ \log\left(\frac{p_\theta(d_{as}\mid s)}{q_\phi(d_{as}\mid s)} \right)\right]\\
&\stackrel{(a)}{=}
\mathbb{E}_{p_\theta}\left[ \log\left(\frac{p_\theta(d_{as}\mid s)}{q_\phi(d_{as})} \right)\right] \\
&= 
\mathbb{E}_{p_\theta}\left[ \log\left(\frac{p_\theta(d_{as}) p_\theta(s \mid d_{as})}{q_\phi(d_{as})p_\theta(s)} \right)\right] \\
&= 
\mathbb{E}_{p_\theta}\left[ \log\left(\frac{p_\theta(d_{as})}{q_\phi(d_{as})}\right) + \log \left(\frac{ p_\theta(s \mid d_{as})}{p_\theta(s)} \right)\right]\\
&= 
\mathbb{E}_{p_\theta}\left[ \log\left(\frac{p_\theta(d_{as})}{q_\phi(d_{as})}\right)\right] +  \mathbb{E}_{p_\theta}\left[\log \left(\frac{p_\theta(s \mid d_{as})}{p_\theta(s)} \right)\right]\\
&= 
D_{KL}(p_\theta(d_{as}) \| q_\phi(d_{as})) + I_\theta(d_{as},s).
\end{align*}
Step (a) holds from the strict fairness constraint, i.e., $I_\phi(d_{as}, s) = 0$, which makes $q_\phi(d_{as}\mid s) = q_\phi(d_{as})$. In addition, the second term $I_\theta(d_{as},s)$ is constant under $p_\theta$. 
Therefore, $D_{KL}(p_\theta(d_{as}) \| q_\phi(d_{as}))$, is minimized when $q_\phi(d_{as}) = p_\theta(d_{as})$. 
This completes our proof. 

\end{proof}

\begin{theorem}
When using Eq \eqref{eq:dec-mix}, the fairness-utility total loss is upper bounded.
Specifically
\begin{align*}
    I_\phi(s, d_{as}) + D_{KL}(p_\theta \| q_\phi) 
    \leq I_{\theta}(d_{as}, s).
\end{align*}
\end{theorem}

\begin{proof}
    
For brevity, we denote $\lambda = \lambda(s, \beta)$.
By definition
\begin{align*}
q_\phi(s, d_{as},d_s)  
&= p_\theta(s) \big(\lambda p_\theta(d_{as}) + (1-\lambda) p_\theta(d_{as} \mid s) \big) p_\theta(d_s \mid d_{as}, s), 
\end{align*}
we have 
\begin{align*}
D_{KL}(p_\theta \|  q_\phi) 
&= 
\mathbb{E}_{p_\theta} \left[ \log\left( \frac{p_\theta(d_{as}, s, d_s)}{q_\phi(d_{as}, s, d_s)} \right) \right] \\
&=
\mathbb{E}_{p_\theta} \left[ \log\left( \frac{p_\theta(s) p_\theta(d_{as} \mid s) p_\theta(d_s  \mid  d_{as}, s)}{p_\theta(s) \left( \lambda p_\theta(d_{as}) + (1 - \lambda) p_\theta(d_{as} \mid s) \right) p_\theta(d_s  \mid  d_{as}, s)} \right) \right] \\
&= 
\mathbb{E}_{p_\theta}\left[ \log\left( \frac{p_\theta(s)}{p_\theta(s)} \right) \right] + \mathbb{E}_{p_\theta} \left[ \log\left( \frac{p_\theta(d_{as} \mid s)}{\lambda p_\theta(d_{as}) + (1 - \lambda) p_\theta(d_{as} \mid s)} \right) \right] + \\
&\quad\quad
\mathbb{E}_{p_\theta} \left[ \log\left( \frac{p_\theta(d_s  \mid  d_{as}, s)}{p_\theta(d_s  \mid  d_{as}, s)} \right) \right] \\
&{=} 
D_{KL}\left( p_\theta(d_{as} \mid s) \| \lambda p_\theta(d_{as}) + (1 - \lambda) p_\theta(d_{as} \mid s) \right)\\
&\stackrel{(a)}{\leq} 
\lambda D_{KL}\left( p_\theta(d_{as} \mid s) \|  p_\theta(d_{as})\right) \\
&= 
\lambda  I_{\theta}(d_{as}, s).
\end{align*}
Step (a) holds from the convexity of KL divergence~\citep{cover1999elements}. 
On the other hand, 
\begin{align*}
I_{\phi}(s, d_{as}) 
&= 
D_{KL}(q_\phi(d_{as} \mid s)\ \| p_\theta(d_{as}))\\
&=  
D_{KL}(\lambda p_\theta(d_{as}) + (1-\lambda) p_\theta(d_{as} \mid s)\| p_\theta(d_{as}))\\
&\stackrel{(a)}{\leq} 
\lambda D_{KL}(p_\theta(d_{as})\| p_\theta(d_{as})) + (1-\lambda) D_{KL}(p_\theta(d_{as} \mid s)\| p_\theta(d_{as}))\\
&=
(1-\lambda) D_{KL}(p_\theta(d_{as} \mid s)\| p_\theta(d_{as}))\\
&= 
(1-\lambda) I_{\theta}(d_{as}, s),
\end{align*}
where step (a) again applies the convexity. 
Put together, 
\begin{align*}
D_{KL}(p_\theta \|  q_\phi) + I_{q_\phi}(d_{as},s) \leq I_{p_\theta}(d_{as}, s).
\end{align*}

This completes our proof. 

\end{proof}

\clearpage
\begin{algorithm}
\caption{Adaptive Inference-Time Debiasing}
\label{alg:adaptive-debiasing}
\begin{algorithmic}[1]
\REQUIRE Pre-trained LLM \(p_\theta\); lightweight MLP \(\lambda(\cdot,\cdot)\); number of iterations \(T\); a set of different hyperparameter \(\{\beta_j\}_{j=1}^M\).
\FOR{\(t=1,\dots,T\)}
    \STATE For each \(\beta_j\), compute 
    \[
    q_\phi(d_{as}\mid s) = \lambda(\beta_j, s)\, p_\theta(d_{as}) + \big(1-\lambda(\beta_j, s)\big)\, p_\theta(d_{as}\mid s).
    \]
    \STATE Evaluate the debiasing objective in Eq.~\eqref{eq:debias-obj} for each \(\beta_j\), average the resulting objectives over all \(\beta_j\)'s, and update \(\) using the averaged objective.
\ENDFOR
\RETURN Trained \(\lambda(\cdot,\cdot)\).
\end{algorithmic}
\end{algorithm}

\begin{algorithm}
\caption{Universal Debiasing Framework with DPO (UDF-DPO)}
\label{alg:udf-dpo}
\begin{algorithmic}[1]
\REQUIRE Pre-trained LLM $p_{\theta}$ (initialized as $q_{\phi}$); number of DPO epochs $T$; reward gap threshold $\delta$; number of samples per epoch $N$
\FOR{$t=1,\dots,T$}
    \STATE \textbf{Step 1:}\textit{ Score each sample}
    \STATE Generate a dataset $\mathcal{D}_{\text{gen}} = \{ d_i \}_{i=1}^N$ from the current model $q_{\phi}$. For each sample $d_i = (s_i, d_{\text{as},i}, d_{\text{s},i}) \in \mathcal{D}_{\text{gen}}$
    \STATE Compute reward 
        \[
            r_i = \log \frac{q_{\phi}(d_{\text{as},i} \mid s_i)}{q_{\phi}(d_{\text{as},i})},
        \]
        where higher reward indicates less bias.
    \STATE
    \STATE \textbf{Step 2:}\textit{ Construct preference pairs}
    \STATE Initialize an empty preference dataset $\mathcal{D}_{\text{pref}} = \emptyset$. For two randomly picked samples $d_i = (s_i, d_{\text{as},i}, d_{\text{s},i})$ and  $d_j = (s_j, d_{\text{as},j}, d_{\text{s},j})$,

        \IF{$|r_i - r_j| > \delta$}
            \IF{$r_i > r_j$}
                \STATE Add preference pair $(y_w = d_i, y_l = d_j)$ to $\mathcal{D}_{\text{pref}}$.
            \ELSE
                \STATE Add preference pair $(y_w = d_j, y_l = d_i)$ to $\mathcal{D}_{\text{pref}}$.
            \ENDIF
        \ENDIF
    \STATE
    \STATE \textbf{Step 3:} \textit{DPO update}
    \STATE Update model parameters $\phi$ using the DPO loss on the preference dataset $\mathcal{D}_{\text{pref}}$.
    \STATE
    \STATE \textbf{(Optional) Step 4:} \textit{Refreshing the samples is implicitly handled by regenerating at the start of the next epoch.}
\ENDFOR
\STATE \RETURN Trained debiased model $q_{\phi}$.
\end{algorithmic}
\end{algorithm}

\begin{table*}[htb!]
\resizebox{\linewidth}{!}{%
\begin{tabular}{r ccccc c ccccc}
\toprule[0.3ex]
& \multicolumn{2}{c}{\textbf{Utility $\uparrow$}} & \multicolumn{3}{c}{\textbf{Bias $\downarrow$}} && \multicolumn{2}{c}{\textbf{Utility $\uparrow$}} & \multicolumn{3}{c}{\textbf{Bias $\downarrow$}} \\
\cmidrule(lr){2-3} \cmidrule(lr){4-6} \cmidrule(lr){8-9} \cmidrule(lr){10-12}
& \textbf{Acc.} & \textbf{AUROC} & \textbf{MI} & \textbf{DP} & \textbf{EO} && \textbf{Acc.} & \textbf{AUROC} & \textbf{MI} & \textbf{DP} & \textbf{EO} \\
\midrule[0.2ex]
\multicolumn{12}{l}{\qquad\qquad \qquad Task 1: Approval–Race \qquad\qquad\qquad\qquad\qquad\qquad\qquad\quad  Task 2: Employment Status–Gender} \\
\midrule[0.2ex]
Original        & 86.13$_{0.31}$ & 88.93$_{1.18}$ & 5.03          & 25.90$_{1.55}$ & 50.90$_{4.35}$ && 96.79$_{0.47}$ & 96.87$_{0.57}$ & 3.93          & 30.31$_{1.06}$ & 66.17$_{8.14}$ \\
Great           & 87.37$_{0.36}$ & 87.05$_{0.62}$ & 3.86$_{0.47}$ & 23.37$_{1.61}$ & 42.16$_{4.80}$ && 95.96$_{0.36}$ & 97.90$_{0.29}$ & 2.65$_{0.07}$ & 28.98$_{0.51}$ & 64.53$_{7.86}$ \\
\midrule[0.2ex]
DECAF-DP        & 85.91$_{1.23}$ & 87.51$_{1.08}$ & \textbf{0.08$_{0.91}$} & \textbf{2.24$_{1.90}$}  & \textbf{4.93$_{1.94}$}  && 83.79$_{0.01}$ & 70.96$_{0.01}$ & 1.70$_{1.09}$ & 15.37$_{4.74}$ & 21.77$_{9.90}$ \\
FairTabGAN      & 82.31$_{2.94}$ & 84.23$_{1.56}$ & 0.14$_{0.07}$ & 4.23$_{1.54}$  & 20.48$_{3.75}$ && 83.65$_{1.09}$ & 73.68$_{1.47}$ & 1.58$_{1.87}$ & 18.32$_{4.97}$ & 20.67$_{1.48}$ \\
FairGAN         & 84.23$_{2.34}$ & 88.42$_{1.29}$ & 0.23$_{0.24}$ & \underline{3.42$_{1.28}$}  & 32.61         && 73.57$_{1.37}$ & 63.49$_{2.68}$ & 0.92$_{4.21}$ & 19.57$_{1.70}$ & 22.12$_{1.24}$ \\
\midrule[0.2ex]
\rowcolor{gray!10}{$\namix$} & & & & & && & & & & \\

$\beta=0.1$     & \textbf{88.82$_{0.80}$} & \textbf{89.54$_{0.66}$} & 1.12$_{0.05}$ & 17.69$_{1.41}$ & 47.70$_{3.90}$ && \textbf{89.23$_{0.39}$} & \textbf{91.60$_{0.30}$} & 0.42$_{0.12}$ & 12.97$_{1.15}$ & 34.98$_{3.11}$ \\
$\beta=1$       & 81.96$_{0.41}$ & 86.35$_{0.17}$ & \underline{0.10$_{0.03}$} & 5.54$_{1.08}$  & 10.90$_{2.54}$ && 71.14$_{0.52}$ & 84.29$_{0.26}$ & 0.68$_{0.05}$ & \textbf{7.88$_{0.88}$}  & \textbf{8.44$_{1.08}$} \\
$\beta=10$      & 81.94$_{0.47}$ & \underline{86.95$_{0.31}$} & 0.29$_{0.09}$ & 7.48$_{2.53}$  & \underline{7.56$_{2.32}$}  && \underline{89.23$_{0.60}$} & \underline{91.21$_{0.41}$} & 0.82$_{0.10}$ & 17.65$_{1.64}$ & 32.75$_{5.39}$ \\
\midrule[0.2ex]
\rowcolor{gray!10}{$\namdpo$} & & & & & && & & & & \\

$\beta=0.1$     & 70.44$_{1.29}$ & 78.78$_{0.77}$ & 0.12$_{0.03}$ & 5.47$_{2.38}$  & 26.93$_{5.02}$ && 85.16$_{0.14}$ & 71.21$_{0.98}$ & \textbf{0.01$_{0.01}$} & \underline{8.52$_{0.31}$}  & 19.15$_{0.49}$ \\
$\beta=1$       & 80.05$_{1.77}$ & 86.24$_{1.27}$ & 0.20$_{0.06}$ & 17.06$_{3.32}$ & 26.43$_{4.97}$ && 85.36$_{0.88}$ & 83.99$_{0.69}$ & \underline{0.05$_{0.06}$} & 9.40$_{0.99}$  & \underline{11.18$_{4.90}$} \\
$\beta=10$      & 73.19$_{0.85}$ & 81.95$_{1.11}$ & 0.17$_{0.11}$ & 10.29$_{2.33}$ & 28.11$_{4.89}$ && 80.17$_{2.37}$ & 84.83$_{0.25}$ & 0.54$_{0.07}$ & 14.78$_{1.98}$ & 14.74$_{1.24}$ \\
\bottomrule[0.3ex]
\end{tabular}
}
\centering
\caption{
Performance on the Credit dataset for two downstream tasks that involve different advantaged–protected feature pairs.
Best results are in \textbf{bold} and second-best results are \underline{underlined}.
\textbf{Baseline methods trained to debias Task 1 remain unfair on Task 2.}
}
\label{tab:credit-combined}
\end{table*}

\begin{table}[htb!]
\definecolor{verylightgray}{gray}{0.9}

\resizebox{0.5\linewidth}{!}{%
\begin{tabular}{r cc c}
\toprule[0.3ex]
 & \multicolumn{2}{c}{\textbf{Utility $\uparrow$}} & \textbf{Bias $\downarrow$} \\ 
\cmidrule(lr){2-3} \cmidrule(lr){4-4}
 & \textbf{Accuracy} & \textbf{RMSE} & \textbf{MI} \\
\cmidrule(lr){2-3} \cmidrule(lr){4-4}
Original        & $-$                & $-$                & 18.56 \\[0.5ex]
\noalign{\vskip 0.2ex}\cdashline{2-4}\noalign{\vskip 0.2ex}
GReaT           & 60.08 $\pm$ 0.42   & 15.12 $\pm$ 0.08   & 18.56 $\pm$ 0.40 \\[0.5ex]
\noalign{\vskip 0.2ex}\cdashline{2-4}\noalign{\vskip 0.2ex}
\rowcolor{gray!10} {\namdpo} & & & \\[0.5ex]
$\beta = 0.1$   & 56.45 $\pm$ 0.28  & 16.67 $\pm$ 0.13   & 15.44 $\pm$ 0.61 \\
$\beta = 1$     & 62.63 $\pm$ 0.60  & 16.41 $\pm$ 0.07   & 15.31 $\pm$ 1.01 \\
$\beta = 10$    & 61.50 $\pm$ 0.32  & 16.94 $\pm$ 0.22   & 15.30 $\pm$ 0.70 \\[0.5ex]
\rowcolor{gray!10} {\namix} & & & \\[0.5ex]
$\beta = 0.1$        & 47.44 $\pm$ 0.22  & 39.87 $\pm$ 41.29  & 15.38 $\pm$ 0.70 \\
$\beta = 1$          & 47.28 $\pm$ 0.65  & 15.91 $\pm$ 0.09   & 14.89 $\pm$ 0.64 \\
$\beta = 10$         & 47.68 $\pm$ 0.16  & 16.08 $\pm$ 0.13   & 15.29 $\pm$ 0.58 \\
\bottomrule[0.3ex]
\end{tabular}
}
\centering
\caption{Data imputation performance on Credit Approval dataset.}
\label{tab:cd-di}
\vspace{-0.2cm}
\end{table}


\begin{table*}[h!]
\centering
\resizebox{\linewidth}{!}{%
\begin{tabular}{r ccccc c ccccc}
\toprule[0.3ex]
& \multicolumn{2}{c}{\textbf{Utility }} & \multicolumn{3}{c}{\textbf{Bias }} && \multicolumn{2}{c}{\textbf{Utility }} & \multicolumn{3}{c}{\textbf{Bias}} \\
\cmidrule(lr){2-3} \cmidrule(lr){4-6} \cmidrule(lr){8-9} \cmidrule(lr){10-12}
& \textbf{Acc.} & \textbf{AUROC} & \textbf{MI} & \textbf{DP} & \textbf{EO} && \textbf{Acc.} & \textbf{AUROC} & \textbf{MI} & \textbf{DP} & \textbf{EO} \\
\midrule[0.2ex]
\multicolumn{12}{l}{\qquad\qquad\qquad\qquad\qquad Task 1: Gender–Income \qquad\qquad\qquad\qquad\qquad Task 2: Race–Education Level}\\
\midrule
\rowcolor{gray!10}UDF-DPO ($\beta=0.1$)& & & & & && & & & &\\
DECAF-DP    & \textbf{0.49$\uparrow$} & 5.10$\downarrow$ & 0.26$\downarrow$ & 0.27$\downarrow$ & 0.24$\downarrow$  
            && \textbf{8.87$\uparrow$} & \textbf{9.69$\uparrow$} & \textbf{0.51$\uparrow$} & \textbf{7.37$\uparrow$} & \textbf{7.79$\uparrow$} \\
TabFairGAN  & 4.15$\downarrow$ & 1.75$\downarrow$ & 0.29$\downarrow$ & \textbf{2.83$\uparrow$} & \textbf{16.64$\uparrow$} 
            && 2.06$\downarrow$ & 6.84$\downarrow$ & \textbf{1.31$\uparrow$} & \textbf{6.17$\uparrow$} & \textbf{4.43$\uparrow$} \\
FairGAN     & \textbf{0.74$\uparrow$} & \textbf{7.32$\uparrow$} & 0.28$\downarrow$ & \textbf{4.89$\uparrow$} & \textbf{7.63$\uparrow$}  
            && \textbf{21.95$\uparrow$} & \textbf{19.85$\uparrow$} & \textbf{0.83$\uparrow$} & \textbf{20.94$\uparrow$} & \textbf{21.88$\uparrow$} \\

\rowcolor{gray!10}UDF-MIX ($\beta=0.1$)& & & & & && & & & &\\
DECAF-DP    & \textbf{6.13$\uparrow$} & 0.40$\downarrow$ & \textbf{0.02$\uparrow$} & 4.87$\downarrow$ & 9.44$\downarrow$  
            && \textbf{8.82$\uparrow$} & \textbf{13.79$\uparrow$} & \textbf{0.43$\uparrow$} & \textbf{5.99$\uparrow$} & \textbf{6.47$\uparrow$} \\
TabFairGAN  & \textbf{1.49$\uparrow$} & \textbf{2.95$\uparrow$} & 0.01$\downarrow$ & 1.77$\downarrow$ & \textbf{7.44$\uparrow$}  
            && 2.11$\downarrow$ & 2.74$\downarrow$ & \textbf{1.23$\uparrow$} & \textbf{4.79$\uparrow$} & \textbf{3.11$\uparrow$} \\
FairGAN     & \textbf{6.38$\uparrow$} & \textbf{12.02$\uparrow$} & 0.00 & \textbf{0.29$\uparrow$} & 1.57$\downarrow$  
            && \textbf{21.90$\uparrow$} & \textbf{23.95$\uparrow$} & \textbf{0.75$\uparrow$} & \textbf{19.56$\uparrow$} & \textbf{20.56$\uparrow$} \\

\midrule

\multicolumn{12}{l}{\qquad\qquad\qquad\qquad\qquad Task 1: Age, MJ, FJ, Gender – $1^{\text{st}}$ Grade \qquad\qquad Task 2: Age, MJ, FJ, Gender – $2^{\text{nd}}$ Grade}\\
\midrule
\rowcolor{gray!10}UDF-DPO ($\beta=0.1$) & & & & & && & & & &\\
DECAF-DP    & \textbf{20.51$\uparrow$} & \textbf{21.54$\uparrow$} & 1.41$\downarrow$ & 3.50$\downarrow$ & 4.93$\downarrow$  
            && \textbf{18.22$\uparrow$} & \textbf{15.62$\uparrow$} & 0.25$\downarrow$ & 0.30$\downarrow$ & 0.64$\downarrow$ \\
TabFairGAN  & \textbf{4.28$\uparrow$} & \textbf{5.01$\uparrow$} & \textbf{1.42$\uparrow$} & 0.62$\downarrow$ & 0.35$\downarrow$  
            && \textbf{2.09$\uparrow$} & \textbf{1.62$\uparrow$} & \textbf{4.03$\uparrow$} & \textbf{2.30$\uparrow$} & \textbf{2.34$\uparrow$} \\
FairGAN     & \textbf{6.28$\uparrow$} & \textbf{3.23$\uparrow$} & \textbf{0.09$\uparrow$} & 0.23$\downarrow$ & \textbf{0.15$\uparrow$}  
            && \textbf{1.22$\uparrow$} & \textbf{3.70$\uparrow$} & \textbf{3.45$\uparrow$} & \textbf{2.68$\uparrow$} & \textbf{2.64$\uparrow$} \\

\rowcolor{gray!10}UDF-MIX ($\beta=0.1$)& & & & & && & & & &\\
DECAF-DP    & \textbf{16.12$\uparrow$} & \textbf{19.21$\uparrow$} & 0.14$\downarrow$ & 2.25$\downarrow$ & 4.20$\downarrow$  
            && \textbf{16.82$\uparrow$} & \textbf{15.90$\uparrow$} & 0.03$\downarrow$ & \textbf{0.08$\uparrow$} & 0.07$\downarrow$ \\
TabFairGAN  & 0.11$\downarrow$ & \textbf{2.68$\uparrow$} & \textbf{2.69$\uparrow$} & \textbf{0.63$\uparrow$} & \textbf{0.38$\uparrow$}  
            && \textbf{0.69$\uparrow$} & \textbf{1.90$\uparrow$} & \textbf{4.25$\uparrow$} & \textbf{2.68$\uparrow$} & \textbf{2.91$\uparrow$} \\
FairGAN     & \textbf{1.89$\uparrow$} & \textbf{0.90$\uparrow$} & \textbf{1.36$\uparrow$} & \textbf{1.02$\uparrow$} & \textbf{0.88$\uparrow$}  
            && 0.18$\downarrow$ & \textbf{3.98$\uparrow$} & \textbf{3.67$\uparrow$} & \textbf{3.06$\uparrow$} & \textbf{3.21$\uparrow$} \\

\midrule
\multicolumn{12}{l}{\qquad\qquad\qquad\qquad\qquad Task 1: Approval–Race \qquad\qquad\quad\qquad\qquad\qquad Task 2: Employment Status–Gender} \\
\midrule
\rowcolor{gray!10}UDF-DPO ($\beta=0.1$)& & & & & && & & & &\\
DECAF-DP    & \textbf{2.91$\uparrow$} & \textbf{2.03$\uparrow$} & 1.04$\downarrow$ & 15.45$\downarrow$ & 42.77$\downarrow$  
            && \textbf{5.44$\uparrow$} & \textbf{20.64$\uparrow$} & \textbf{1.28$\uparrow$} & \textbf{2.40$\uparrow$} & 13.21$\downarrow$ \\
TabFairGAN  & \textbf{6.51$\uparrow$} & \textbf{5.31$\uparrow$} & 0.98$\downarrow$ & 13.46$\downarrow$ & 27.22$\downarrow$  
            && \textbf{5.58$\uparrow$} & \textbf{17.92$\uparrow$} & \textbf{1.16$\uparrow$} & \textbf{5.35$\uparrow$} & 14.31$\downarrow$ \\
FairGAN     & \textbf{4.59$\uparrow$} & \textbf{1.12$\uparrow$} & 0.89$\downarrow$ & 14.27$\downarrow$ & 15.09$\downarrow$  
            && \textbf{15.66$\uparrow$} & \textbf{28.11$\uparrow$} & \textbf{0.50$\uparrow$} & \textbf{6.60$\uparrow$} & 12.86$\downarrow$ \\

\rowcolor{gray!10}UDF-MIX ($\beta=0.1$) & & & & & && & & & &\\
DECAF-DP    & 15.47$\downarrow$ & 8.73$\downarrow$ & 0.04$\downarrow$ & 3.23$\downarrow$ & 22.00$\downarrow$  
            && \textbf{1.37$\uparrow$} & \textbf{0.25$\uparrow$} & \textbf{1.69$\uparrow$} & \textbf{6.85$\uparrow$} & \textbf{2.62$\uparrow$} \\
TabFairGAN  & 11.87$\downarrow$ & 5.45$\downarrow$ & \textbf{0.02$\uparrow$} & 1.24$\downarrow$ & 6.45$\downarrow$  
            && \textbf{1.51$\uparrow$} & 2.47$\downarrow$ & \textbf{1.57$\uparrow$} & \textbf{9.80$\uparrow$} & \textbf{1.52$\uparrow$} \\
FairGAN     & 13.79$\downarrow$ & 9.64$\downarrow$ & \textbf{0.11$\uparrow$} & 2.05$\downarrow$ & \textbf{5.68$\uparrow$}  
            && \textbf{11.59$\uparrow$} & \textbf{7.72$\uparrow$} & \textbf{0.91$\uparrow$} & \textbf{11.05$\uparrow$} & \textbf{2.97$\uparrow$} \\
\bottomrule[0.3ex]
\end{tabular}
}
\caption{Each row represents the improvements of our methods with $\beta=1.0$ over the baselines on the Adult dataset (upper), Student Performance (middle), and Credit dataset (lower). The MI is calculated as the absolute difference because its values are small, while the other metrics are calculated in terms of percentage. Improvements are highlighted in bold with an upward arrow. Note that only Task 1's target and protected features are revealed to task-specific baselines during training, whereas our methods debias all potential downstream tasks simultaneously.}
\label{tab:tradeoff}
\end{table*}

\twocolumn[
  \section{Additional Experiment Results and Details}\label{app:add-exp}
]

\myparagraph{Hardware and Implmentation packages. } We use NVIDIA RTX A6000 for all the experiments and utilize the TRL - Transformer Reinforcement Learning
to implement DPO. 

\myparagraph{Convergence. } UDF-DPO and UDF-MIX are trained with 5 and 8 epochs, which are typical convergence points.

\myparagraph{Additional results on diverse datasets. } The table~\ref{tab:credit-combined} additional experiments on the credit dataset for downstream tasks, and table~\ref{tab:cd-di} contains results for data imputation results. The experiments on the Credit dataset further validate our universal debiasing framework, with Task 1 targeting Approval–Race and Task 2 targeting Employment Status–Gender. While models trained on the original data and the backbone generator, GReaT, show significant bias, our proposed methods, UDF-DPO and UDF-MIX, demonstrate strong universal performance by reducing bias across both tasks simultaneously. In contrast, task-specific baselines like DECAF-DP, which are trained only on Task 1, remain unfair when evaluated on Task 2, highlighting the limitations of pairwise debiasing that our framework overcomes. Furthermore, our methods extend their effectiveness to data imputation tasks, as shown in the results from Table 5. In this setting, both UDF-DPO and UDF-MIX achieve lower Mutual Information, which indicates a more fair generation.

\myparagraph{Trade-off analysis. }
In Table~\ref{tab:tradeoff}, our \emph{universal debiasing} methods demonstrate a better balance between the data utility and fairness on \emph{Task~2} across datasets, achieving utility gains of up to \({21.95}\) Acc and \({23.95}\) AUROC (Adult), and as high as \({28.11}\) AUROC (Credit), while simultaneously improving fairness by as much as \(1.31\) MI, \({20.94}\) DP, and \({21.88}\) EO. In contrast, on \emph{Task~1} our gains are generally modest, often within \(6.4\) Acc and \(12.0\) AUROC on Adult, reflecting the expected trade-off when baselines are specialized for the pairwise debiasing. The pattern supports our \emph{universal debiasing} objective: unlike pairwise baselines that only debias a single advantaged-protected pair, our methods focus on group-wise independence that achieves debiasing to unseen pairs (Task~2).

\end{document}